\documentclass[AMA,STIX1COL]{WileyNJD-v2}

\usepackage[utf8]{inputenc}
\usepackage{xcolor}
\usepackage{color,soul}
\usepackage{graphicx}
\usepackage{amsmath}
\usepackage[version=4]{mhchem}
\usepackage{siunitx}
\usepackage{longtable,tabularx}
\usepackage{caption}
\usepackage{subcaption}
\articletype{Article Type}%

\received{26 April 2016}
\revised{6 June 2016}
\accepted{6 June 2016}

\begin{document}

\title{Rapid Uncertainty Propagation and Chance-Constrained Path Planning for Small Unmanned Aerial Vehicles}

\author[1]{Andrew W. Berning Jr.*}

\author[2]{Anouck Girard}

\author[3]{Ilya Kolmanovsky}

\author[4]{Sarah N. D'Souza}

\authormark{Berning \textsc{et al}}

\address[1]{\orgdiv{Research Assistant}, \orgname{Department of Aerospace Engineering, University of Michigan}, \orgaddress{\state{MI}, \country{USA}}}

\address[2]{\orgdiv{Associate Professor}, \orgname{Department of Aerospace Engineering, University of Michigan}, \orgaddress{\state{MI}, \country{USA}}}

\address[3]{\orgdiv{Professor}, \orgname{Department of Aerospace Engineering, University of Michigan}, \orgaddress{\state{MI}, \country{USA}}}

\address[4]{\orgdiv{Aerospace Flight Systems Engineer}, \orgname{NASA Ames Research Center}, \orgaddress{\state{CA}, \country{USA}}}

\corres{\email{awbe@umich.edu}}

\abstract[Summary]{With the number of small Unmanned Aircraft Systems (sUAS) in the national airspace projected to increase in the next few years, there is growing interest in a traffic management system capable of handling the demands of this aviation sector. It is expected that such a system will involve trajectory prediction, uncertainty propagation, and path planning algorithms. In this work, we use linear covariance propagation  in combination with a quadratic programming-based collision detection algorithm to rapidly validate declared flight plans. Additionally, these algorithms are combined with a Dynamic, Informed RRT* algorithm, resulting in a computationally efficient algorithm for chance-constrained path planning. Detailed numerical examples for both fixed-wing and quadrotor sUAS models are presented. }


\maketitle

\section{Introduction} \label{Introduction}
\subsection{Motivation}

There were 110,604 registered small Unmanned Aircraft Systems (sUAS) in the United States at the end of 2017, and that number is expected to quadruple by 2022 \cite{schaufele2018faa}. There has been great interest, accordingly, in a UAS Traffic Management (UTM) system to handle the demands of this rapidly growing aviation sector \cite{kopardekar2014unmanned,kopardekar2016unmanned,dill2016safeguard}. In at least one potential design of such a system, real-time communication channels exist between a central computation platform and the sUAS. This would provide the vehicles access to valuable computation resources and information about nearby vehicles, terrain, and atmospheric conditions, allowing for safe and efficient route planning.

One of the responsibilities of such a UTM system will likely be to apply a risk-based approach where geographical needs and use cases determine the airspace performance requirements \cite{dsou:17cp}. Addressing this will require a sUAS trajectory prediction model that validates vehicle flight performance and allows for UTM 
to determine whether or not the vehicle can operate in the airspace given real-time information about wind, other vehicles, and/or obstacles. Given a large number of vehicles predicted to be in operation, this trajectory prediction model must accommodate multiple vehicle types and airspace environments (wind, terrain, etc.). Another challenge in this setting is that the trajectory could depend on proprietary  information such as control systems, methods and gain tuning specific to a particular vehicle. Implementing a system that is reliant on such information 
could be prohibitive due to constant modifications that would be required to accommodate diverse and evolving control laws and due to the potential legal barriers to acquiring proprietary information from all operators.

In the context of UTM operations, where varied vehicle types and uncertainties are expected, there is a need 
%for a trajectory simulation that trades a vehicle-centric approach with a systems level approach to 
for a model that represents expected vehicle performance while accounting for the uncertainty in the context of operational environments. The motivation for the work detailed in this paper is to develop computationally efficient trajectory validation and  planning algorithms that utilize uncertainty quantification (UQ) and propagation techniques 
%for wind and aerodynamic uncertainty propagation 
to provide an assessment of the vehicle performance and risk
of violating constraints. % volume of operation. 

\subsection{Problem Statement}

The work described in this paper focuses on two separate but related problems: rapid uncertainty propagation for trajectory validation and chance-constrained path planning. 

\begin{figure}[htbp!]
	\centering
	\includegraphics[width=0.5\linewidth]{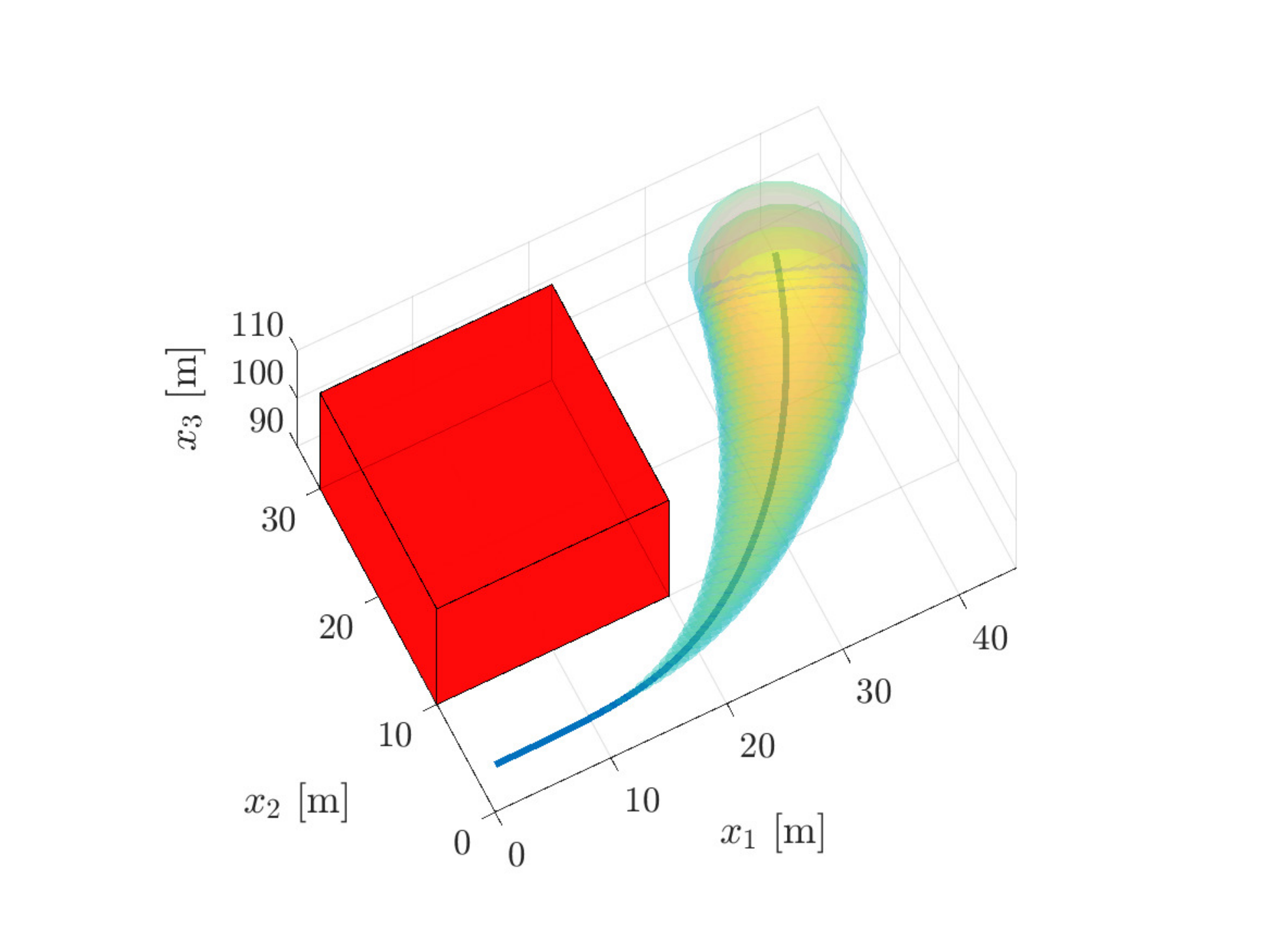}
	\caption{Nominal trajectory (dotted), probabilistic trajectory tube  in the presence of uncertainties and an obstacle (red).}
	\label{fig:volume_of_operation}
\end{figure}

For the former, we seek an algorithm that takes the vehicle's dynamics, initial state, parameters, and desired trajectory as inputs, and outputs some quantification of the uncertainty associated with the vehicle's state trajectory over the specified flight horizon, as well as an assessment of whether the probability of trajectories violating constraints is sufficiently low at any given time instant. For the latter, we are interested in the ability to re-plan or repair the trajectory if it is found that the vehicle's probabilistic trajectory tube intersects with a keep-out zone. Such a tube and constraint are illustrated in Figure \ref{fig:volume_of_operation}.

\subsection{Literature Review}

The source of uncertainty considered in this work is atmospheric turbulence and stochastic wind gusts, which affect the response \cite{richardson2013envelopes} and safety \cite{belcastro2010aircraft,FAA2004accidents} of aircraft. These effects are particularly pronounced in relatively lightweight sUAS. Hoblit \cite{hoblit1988gust} provides a detailed description of discrete and continuous gust models and Richardson \cite{richardson2013quantifying} provides a thorough review of the modeling  techniques involved. The Dryden and Von K\'arm\'an models are the two most common gust/turbulence models \cite{liepmann1952application,de1938statistical,von1948progress,von1951statistical,diederich1957effect}, which are used in both Federal Aviation Administration (FAA) and military specifications. Both consider the linear and angular velocity components of the gusts to be varying stochastic processes and then specify those components' Power Spectral Density (PSD). The Dryden model utilizes rational PSDs, while the Von K\'arm\'an model uses irrational PSDs. The latter matches experimental gust observations more closely than the former, but its use of irrational spectral densities prevents its spectral factorization from being exactly expressed. 

We also consider the problem of propagating the vehicle's state uncertainty subject to the aforementioned wind gust disturbances. Given an initial probability distribution of a state, the objective of a UQ algorithm is to obtain a characterization of the state's probability distribution at a future instant in time. For this work, we require the algorithm to compute the uncertainty for all time instants between the initial and final times. The conceptually simple solution is through a Monte Carlo (MC) simulation, but the computational intensity of this method limits its usefulness for our application \cite{junkins1996non,sabol2013nonlinear}. Linearization techniques suffer from diminished accuracy for highly nonlinear systems or for long time horizons, but their simplicity and high computational efficiency make them well-suited for real-time trajectory optimization or path planning \cite{geller2006linear,geller2009event}. Other nonlinear UQ methods include unscented transformation (UT) \cite{julier1995new,julier2000new}, polynomial chaos (PC) expansions \cite{wiener1938homogeneous}, and Gaussian mixture models (GMM) \cite{terejanu2008uncertainty,demars2013entropy,vittaldev2016space}. A thorough survey of many other UQ methods is provided by Luo \cite{luo2017review}, though none of these were deemed appropriate for our use due to our models' large number of states, the presence of stochastic inputs, and the need for rapid computations. For that reason, the linear covariance propagation method is 
an attractive choice for our problem. The ease of incorporating exogenous disturbances and computing a complete time history of the covariance further distinguish it from the other UQ methods. 

There are a wide array of solution strategies for planning a vehicle's path subject to uncertainty, including convex programming \cite{blackmore2011chance}, mixed integer linear programming \cite{da2019collision}, graph search \cite{bry2011rapidly}, fast marching trees \cite{janson2018monte}, and probabilistic roadmaps \cite{ding2013strategic}. One of the more popular approaches \cite{luders2010chance,pairet2018uncertainty,axelrod2018provably,kothari2013probabilistically,borowski2012evaluation} utilizes a class of stochastic search algorithms called Rapidly-exploring Random Trees (RRTs) \cite{lavalle1998rapidly}. These algorithms are well suited for real-time implementation \cite{pairet2018uncertainty} and are sampling-based, so they scale well with problem size, but only offer a probabilistic guarantee of completeness. A comparison of sequential quadratic programming, genetic algorithms, and RRT is provided in \cite{borowski2012evaluation}. Extensions to RRT such as RRT*\cite{karaman2011sampling}, Informed RRT*\cite{gammell2014informed}, and Dynamic RRT*\cite{adiyatov2017novel}, improve optimality of the solution, increase sampling efficiency, and allow for dynamically changing constraints, respectively. These three RRT extensions are utilized in this work.

\subsection{Original Contributions}

The main contribution of this paper is the amalgamation of existing Dynamic RRT* and Informed RRT* algorithms, and the addition of an obstacle buffer resizing technique, resulting in a single algorithm that allows computationally efficient, chance-constrained path planning for sUAS. In particular, this contribution addresses a dilemma in chance-constrained path planning under uncertainty: trajectory re-planning changes the outcome of the covariance propagation, which, when obstacles or exclusion zones are involved, may require further re-planning.

This paper builds upon previous work \cite{berning2018Rapid} that considered rapid uncertainty propagation, collision detection, and trajectory optimization for fixed-wing, 2D longitudinal flight dynamics. The numerical examples presented here utilize both fixed-wing and quadrotor sUAS in full 3D flight in a non-static atmosphere with inner loop and outer loop controllers in closed form. Other contributions in support of the path planning algorithm are linear covariance propagation of the uncertainty in initial states and exogenous disturbances for three dimensional vehicle motion, and a quadratic programming-based approach to 3D collision detection. 

\section{Modeling} \label{sec:modeling}

The purpose of this paper is to present and illustrate with simulations a procedure for rapid uncertainty propagation and chance constrained path planning that is broadly applicable in the sUAS setting in terms of models and uncertainties assumed. This section introduces the two systems used for numerical examples in this work: a quadrotor sUAS model and a fixed-wing sUAS model. 

The quadrotor sUAS is modeled as a double integrator with quadratic aerodynamic drag, constant drag coefficient, dynamic extension controller, and a Dryden-based wind disturbance. The full model can be found in Appendix \ref{app:quad_model}. The fixed-wing sUAS model is based on a 6-state model for longitudinal and lateral flight in a moving atmosphere, with inner and outer-loop controllers added and a Dryden wind gust model. The full model can be found in Appendix \ref{app:fixed_model}. 

The above models given by
Eqs. \eqref{eq:quad_dr} -- \eqref{eq:quad_wi} for the quadrotor or Eqs. \eqref{fixed_dx} -- \eqref{dgamma} for the fixed-wing aircraft can be viewed 
as a single system of differential equations of the form: 

\begin{equation}
\dot{X} = f\big (X,X_{des}(t),n(t),\theta \big), \label{eq:dX}
\end{equation}

\noindent where the state is $X = [{r}, {V}_0, \eta_{x1}, \eta_{x2}, \eta_{x3}]^\top$ for the quadrotor model or $X = [x,y,h,V,\psi,\gamma,T,V_{des},\psi_{des},\eta_u,\eta_w,\eta_v ]^\top$ for the fixed-wing model, $\theta$ is the set of vehicle, environmental, and control parameters, and $n$ is the white noise input.

\section{Technical Approach}

\subsection{Covariance Propagation} \label{sec:cov_prop}

For the uncertainty propagation analysis, we linearize the sUAS model in Eq. \eqref{eq:dX}. The linearized model has the following form: 

\begin{align}
\delta\dot{X}=&~A(t)\delta X(t)+B_n(t)n(t), \label{eq:linearization}\\
A(t) = &~\frac{\partial f}{\partial X}\Biggr|_{\substack{X=\bar{X}(t) \\ n=0 \\ X_{des}(t)}}, \\
B_n(t) = &~\frac{\partial f}{\partial n}\Biggr|_{\substack{X=\bar{X}(t) \\ n=0\\ X_{des}(t)}}.
\end{align}

Here, $\bar{X}(t)$ is the nominal trajectory obtained by propagating Eq.~\eqref{eq:dX} subject to $n=0$ and specified time history of $X_{des}$. Physically, this corresponds to the nominal trajectory that the sUAS would fly under no-wind conditions. 

Under the assumption that $n(t)$ is a standard white noise process, the state covariance matrix, $P$, satisfies the Lyapunov differential equation \cite{kabamba2014fundamentals}:
\begin{align}
\dot{P}(t)=&~A(t)P(t)+P(t)A^{\top} (t)+B_n(t)B^{\top}_n(t). \label{eq:LinCov}
\end{align}

Note that \eqref{eq:LinCov} can be solved using any standard ODE solver. The initial condition for $P$ is the zero matrix if the vehicle's initial state is known deterministically, or may be chosen as a diagonal matrix of state covariances, otherwise. 

Now, we can use the nominal trajectory $\bar{X}(t)$, covariance matrix $P(t)$, and a Gaussian distribution density function to estimate the probability that the vehicle is contained in a specified set at time $t$. For the quadrotor model: given the mean $\bar{r} (t)= [\bar{x}_1(t),\bar{x}_2(t),\bar{x}_3(t)]^{\top}$ and the block of the covariance matrix corresponding to these states, $\Sigma_\omega (t)=P_{(1:3,1:3)}(t)$, we can build a set from the definition of a probability ellipsoid \cite{malyshev1992optimization}:

\begin{equation} \label{eq:prob_ellipse}
Prob~[\omega \in \mathbb{R}^3:(\omega - \bar{r}(t))^{\top}\Sigma^{-1}_\omega(t) (\omega - \bar{r}(t))\leq c^2]=\beta, 
\end{equation} 

\noindent where $\beta \in [0,1]$ is a prescribed probability level and $c$ is solved for using the three degree-of-freedom chi-squared distribution\cite{lancaster1969chi}. Thus the vehicle has a $\beta$ probability of being within the set \eqref{eq:prob_ellipse} at time $t$.

For verification purposes, Eq.~\eqref{eq:dX} is also solved numerically in a series of MC simulations. For these computations, the white Gaussian noise is computed using MATLAB's $randn()$ function, scaled by the inverse square root of the integration time step, as detailed in \cite{hanson2007applied}. A comparison between the linear covariance propagation and the covariance matrix estimated from the MC runs is shown in Section \ref{results}. 

\subsection{Collision Detection}

\begin{figure}
	\centering
	\begin{subfigure}{.32\linewidth}
		\centering
		\includegraphics[width=\linewidth]{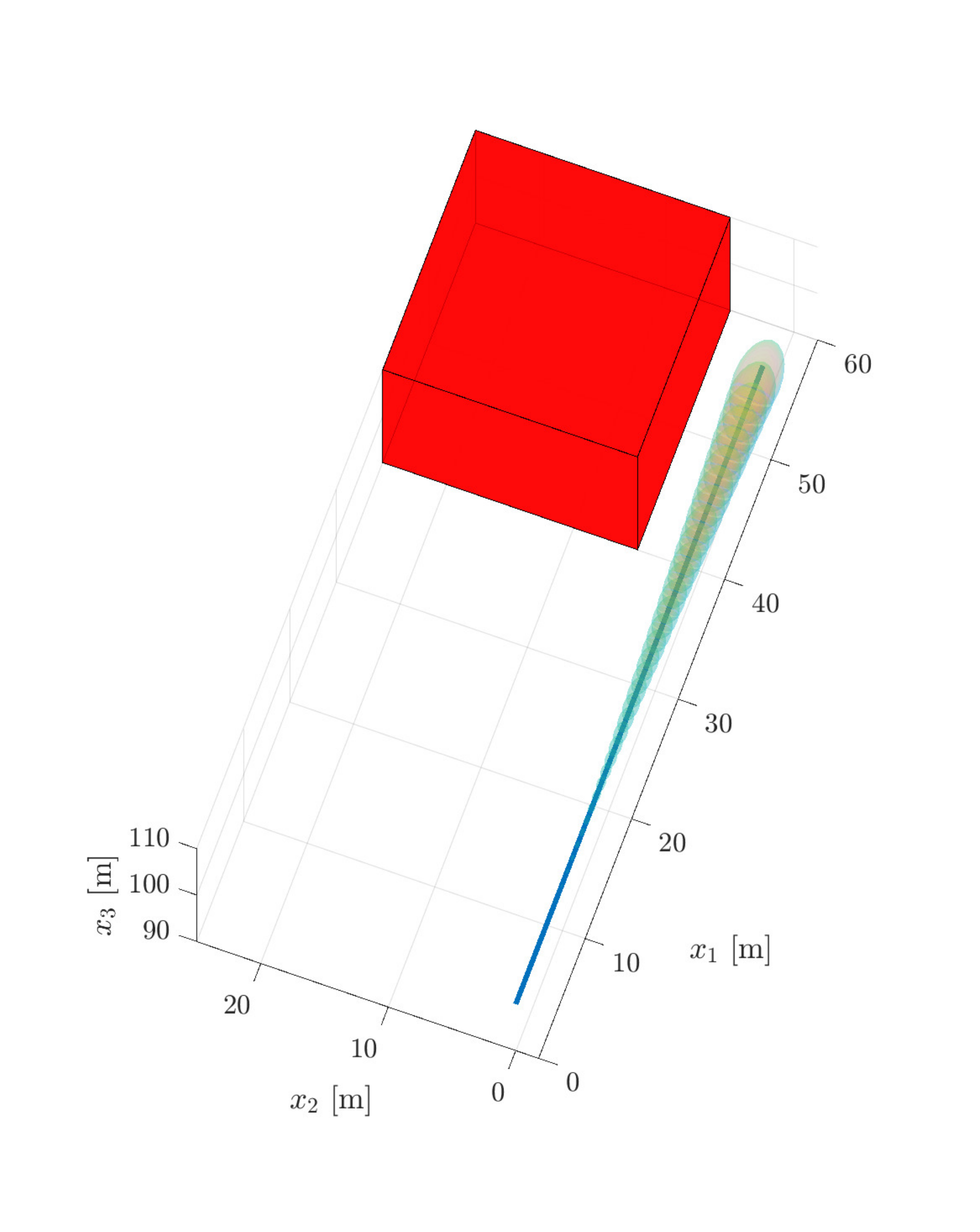}
		\caption{$\beta = 0.5$}
	\end{subfigure} \hfill
	\begin{subfigure}{.32\linewidth}
		\centering
		\includegraphics[width=\linewidth]{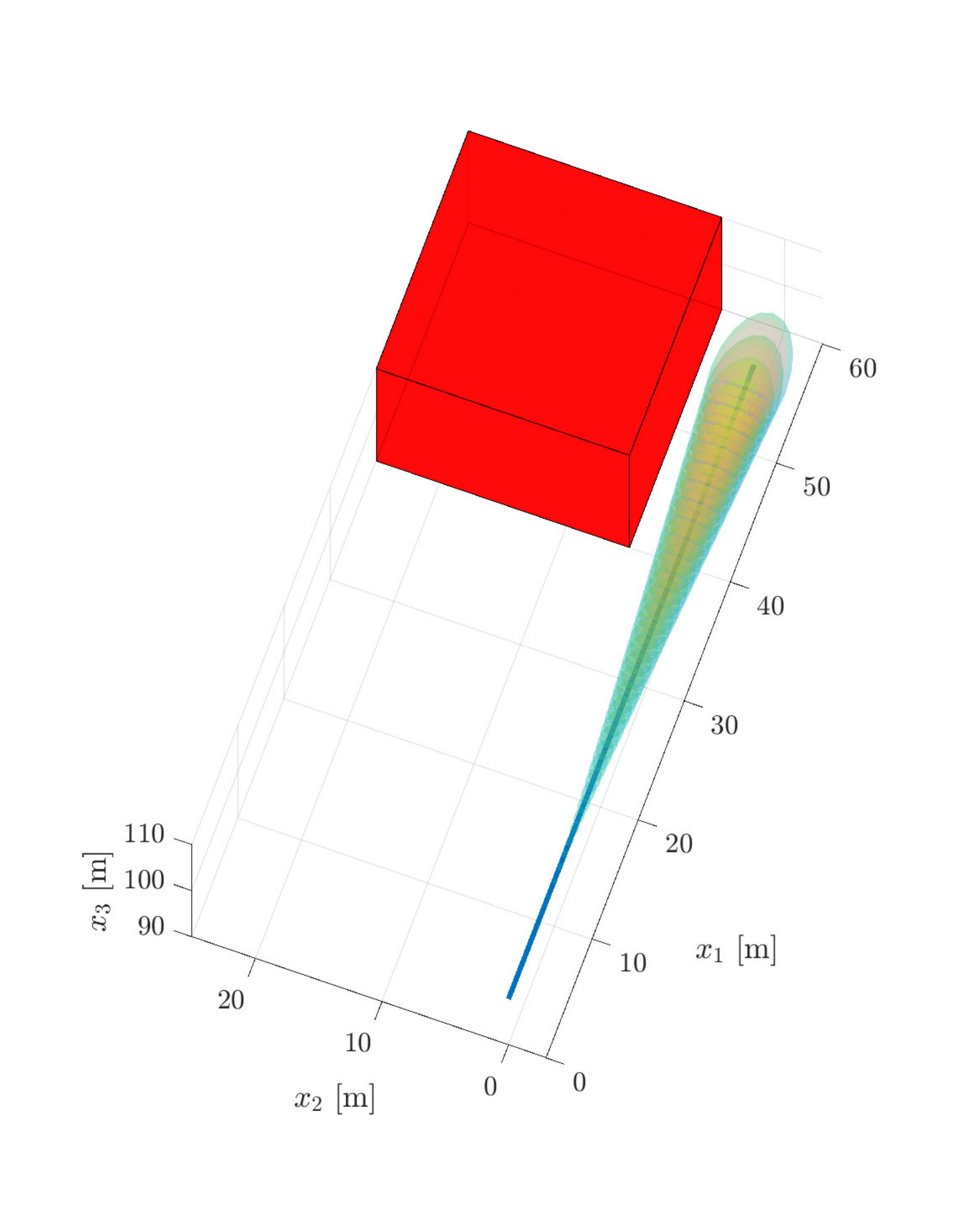}
		\caption{$\beta = 0.9$}
	\end{subfigure} \hfill
	\begin{subfigure}{.32\linewidth}
		\centering
		\includegraphics[width=\linewidth]{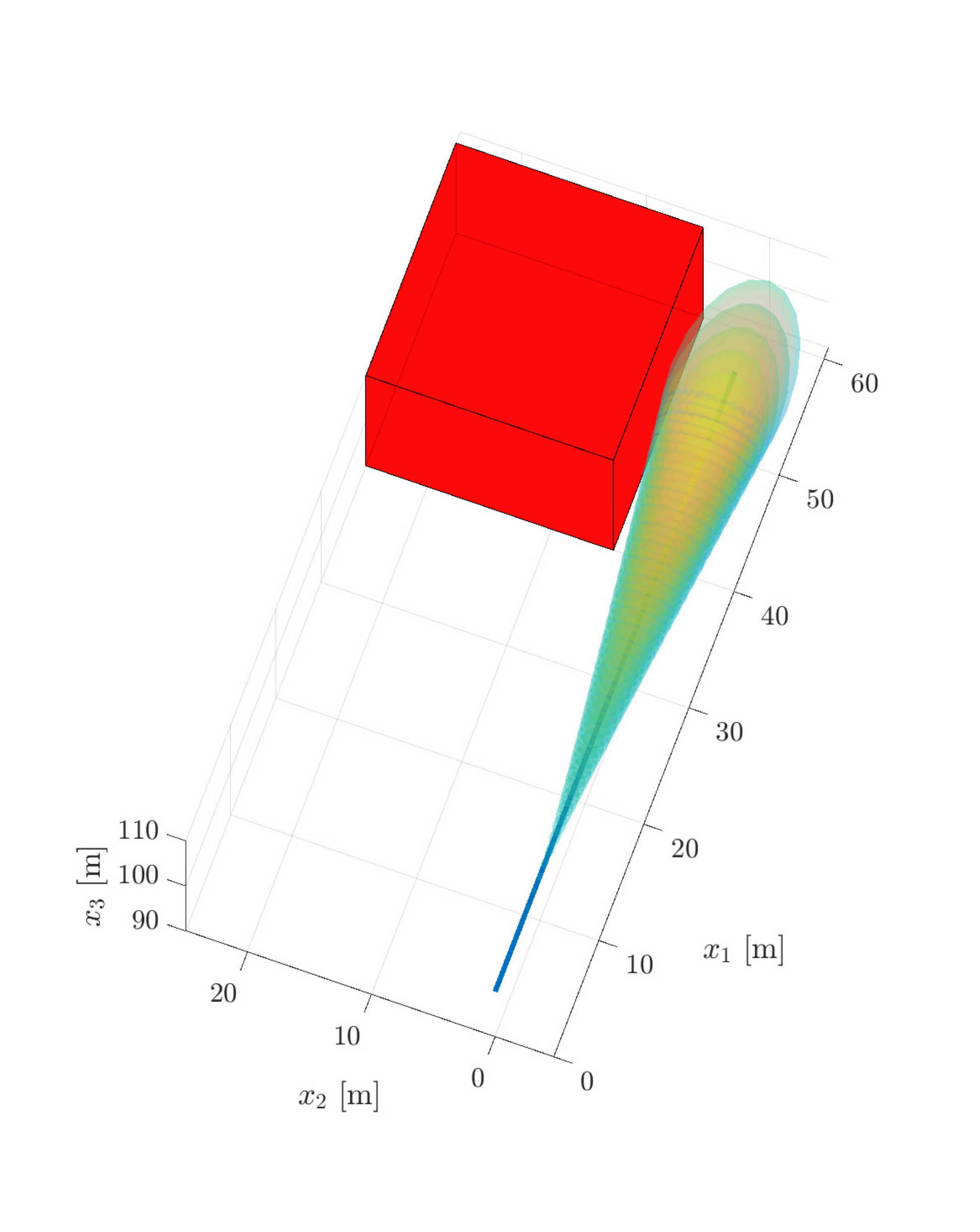}
		\caption{$\beta = 0.999$}
	\end{subfigure} \hfill
	\caption{ Collision scenario for varying values of $\beta$.}
	\label{fig:collision_ex}
\end{figure}

For trajectory validation purposes, it is desirable to have a collision detection algorithm that can quickly determine whether or not the trajectory tube formed by the set of probability ellipsoids defined in Eqn. \eqref{eq:prob_ellipse} intersects with an obstacle, represented in this work as a cuboid, as illustrated in Figure \ref{fig:collision_ex}.

The ellipsoid-cuboid intersection at time instant $t$ is formulated as a quadratic programming (QP) problem that can be solved by a standard QP solver:
\begin{align}
&\min_z ~  (z - \bar{r}(t))^{\top}\Sigma^{-1}_\omega(t) (z - \bar{r}(t)) \label{eq:QP}\\ 
&s.t. \nonumber \\
& A_Oz\leq b_O . \nonumber
\end{align}

Here, $z$, the optimization parameter, is a concatenation of $x_1$, $x_2$ and $x_3$ coordinates: $z=[x_1, x_2, x_3]^\top$ and $A_Oz\leq b_O$ are the linear inequalities that represent the cuboid obstacle. Thus we are solving for the point $z$ that belongs to the smallest ellipsoid centered at $\bar{r}(t)$ that still touches the cuboid obstacle defined by $A_Oz\leq b_O$. Let $z^*$ denote the solution to this QP and let $c^{*2}(t) = (z^* - \bar{r}(t))^{\top}\Sigma^{-1}_\omega(t) (z^* - \bar{r}(t))$. If $c^{*2}(t) \geq c^2$, then there is no intersection between the ellipsoid and cuboid at time $t$, and the chance constraint is not violated.

For the numerical implementation of this collision detection scheme, the frequency at which collisions are checked is an algorithm tuning parameter, and \eqref{eq:QP} is only solved after it has been determined that the overbounding spheres of the ellipsoid and cuboid intersect. The latter check is computationaly very simple. 

\subsection{Path Planning} \label{sec:path_planning}

In our prototypical UTM scenario mentioned in Section \ref{Introduction}, the covariance propagation and tube generation algorithms discussed in Section~\ref{sec:cov_prop} inform the functionality to validate aircraft trajectories. If a desired flight plan is found to be unsafe due to possible collisions with obstacles, the UTM system should return a safe flight path for the UAS to fly. Here, the UAS is operating in a densely populated urban environment with many no-fly zones or obstacles. For scenarios such as this, with a large number of nonconvex obstacle constraints, a Rapidly-expanding Random Tree (RRT) algorithm is useful. 

This section lays out the modifications made to the standard 2D RRT algorithm \cite{lavalle1998rapidly} for the purpose of generating a desired constant-altitude flight path $X_{des}$. Note that the path planning is handled in two dimensions, under the assumption that the UAS operates in an urban environment at a constant desired altitude, but all tube generation and collision checks are three-dimensional.

\subsubsection{Informed Subset}

By their nature, RRT* algorithms find optimal paths between the initial state and every state in the search space, regardless of what final state the user is actually interested in. This is an inefficient use of computational resources that the Informed variant \cite{gammell2014informed} attempts to address. By only sampling within an ellipsoidal subset and excluding regions of the sample space that cannot possibly improve the solution, the algorithm makes a better-informed decision about where to look for more optimal paths. 

This informed subset is defined as follows:

\begin{align}
Q_{\hat{f}} = \big \{ q\in Q ~\big| ~~\lVert x_{start} - q \rVert + \lVert q- x_{goal} \rVert \leq c_{best} \big \} \label{eq:sample_ellipse}
\end{align}

\noindent where $c_{best}$ is the current best solution cost, $Q$ is the global sample space, $x_{start}$ is the initial state and $x_{goal}$ is the desired final state.

This Informed variant is outlined in Algorithm \ref{alg:IRRTs} from \cite{gammell2014informed}, where $\mathcal{T}$ is the current set of nodes, $r_w$ is the rewiring radius, and $N$ is the total number of iterations. Note that the subroutine $SampleEllipse(x_{start},x_{goal},c_{best})$ returns a randomly sampled point that is contained within the ellipsoid defined in \eqref{eq:sample_ellipse}, and lines \ref{alg:rewire_start}--\ref{alg:rewire_end} in Algorithm \ref{alg:add_node} include the rewiring steps integral to the RRT$^*$ algorithm variant.

\begin{algorithm}
	\caption{$\mathcal{T} = $ Informed RRT$^*$($x_0, x_f$)}
	\label{alg:IRRTs}
	\begin{algorithmic} [1]
		\State $\mathcal{T}_0 \leftarrow x_0 $
		\State $c_{best} \leftarrow \infty $
		\For{$k = 1 \dots N$}
		\State $\mathcal{T}_{k+1} = AddNode(\mathcal{T}_k)$
		\EndFor \\
		\Return{$\mathcal{T}_N$}
	\end{algorithmic}
\end{algorithm}

\begin{algorithm}
	\caption{$\mathcal{T}_{k+1} = AddNode(\mathcal{T}_k)$}
	\label{alg:add_node}
	\begin{algorithmic} [1]
		\State $c_{best} \leftarrow min_{x_{xoln}\in X_{soln}}\{Cost(x_{soln})\}$
		\State $x_{rand} \leftarrow SampleEllipse(x_{start},x_{goal},c_{best})$;
		\State $x_{nearest} \leftarrow Nearest(\mathcal{T},x_{rand})$;
		\State $x_{new} \leftarrow Steer(x_{nearest},x_{rand})$;
		\If{$NoCollision(x_{nearest},x_{new})$}
		\State $\mathcal{T} \leftarrow AddNode(\mathcal{T},x_{new})$;
		\State $X_{near} \leftarrow Near(\mathcal{T},x_{new},r_w)$; \label{alg:rewire_start}
		\State $x_{min} \leftarrow x_{nearest}$;
		\State $c_{min} \leftarrow Cost(x_{min})+ || coordinates(x_{nearest} - coordinates(x_{new}) || $
		\For{$\forall x_{near} \in X_{nearest}$}
		\State $c_{new} \leftarrow Cost(x_{near})+ || coordinates(x_{near} - coordinates(x_{new}) ||$
		\If{$c_{new} < c_{min}$}
		\If{$NoCollision(x_{nearest},x_{new})$}
		\State $x_{min} \leftarrow x_{near}$
		\State $c_{min} \leftarrow c_{new}$
		\EndIf 
		\EndIf 
		\EndFor \label{alg:rewire_end}
		\State $Parent(x_{new}) \leftarrow x_{min}$
		\EndIf \\
		\Return{$\mathcal{T}$}
	\end{algorithmic}
\end{algorithm}

\subsubsection{Chance Constraints and Dynamic Extension}

If the solution of Algorithm \ref{alg:IRRTs} was passed to the tube generation algorithm as $X_{des}$, there would be no guarantee that the tube generated would not intersect the obstacles. The solution to a path planning problem with obstacles often passes very closely to the obstacle, where any nonzero uncertainty in the state would produce a tube that would intersect the obstacle. Thus a buffer region is added to the obstacles and adjusted periodically in a manner informed by full covariance propagation. 

\begin{figure}[htbp!]
	\centering
	\includegraphics[width=0.5\linewidth]{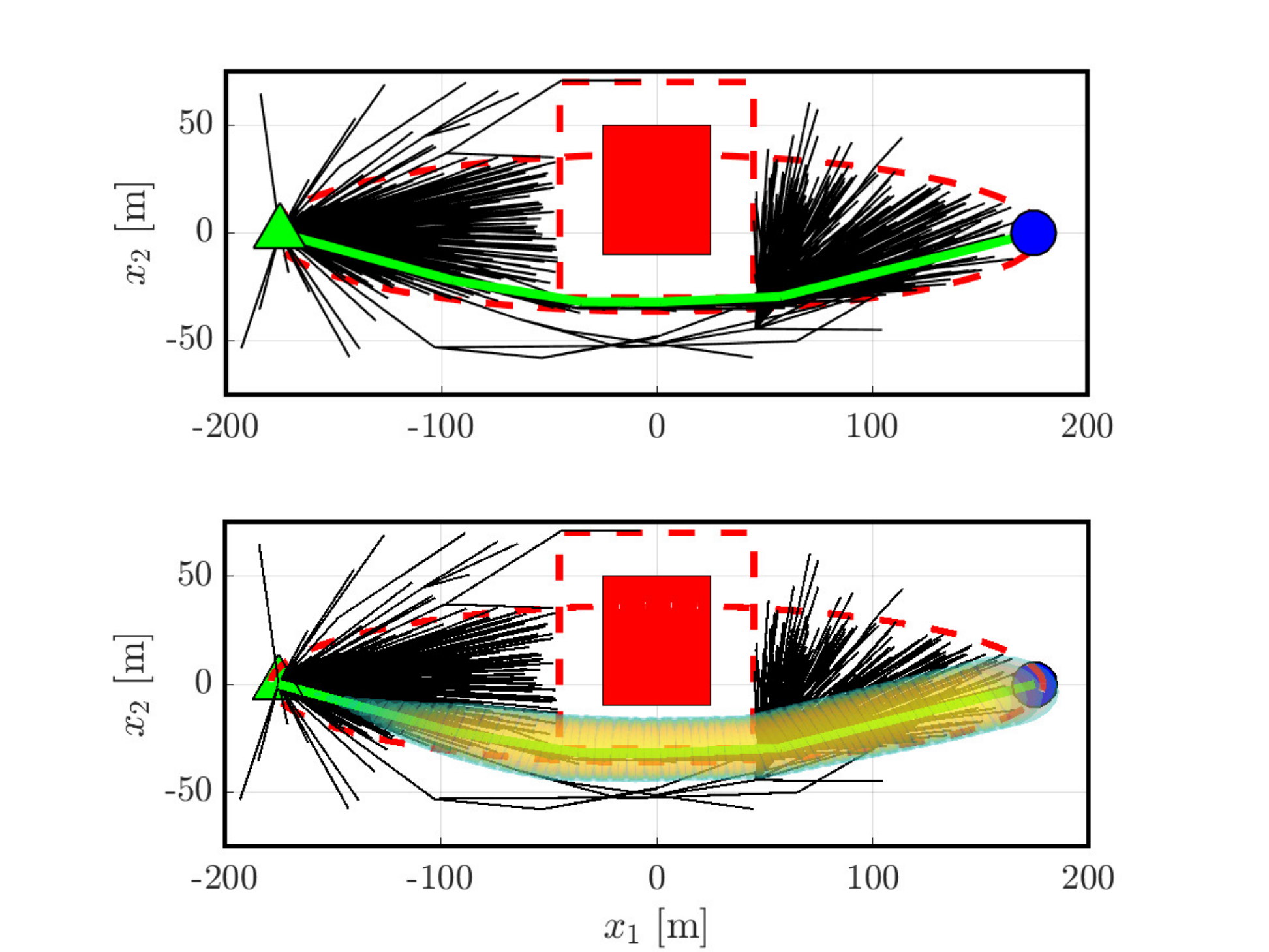}
	\caption{Results of Algorithm \ref{alg:IRRTs} with a buffered obstacle.}
	\label{fig:buffer_example}
\end{figure}

This buffer can be seen in Figure \ref{fig:buffer_example}, which represents a single outer-loop iteration in which the buffer is held constant. Here, the green triangle is the starting point $x_{start}$, the blue circle is the end point $x_{goal}$, the green line is the current best solution, the dashed red ellipse is the current informed subset as defined in Eqn. \eqref{eq:sample_ellipse}, and the black lines represent the total set of sampled nodes and connecting vertices. The solid red rectangle is the actual obstacle, while the dashed red rectangle is the buffered obstacle that is actually passed to Algorithm \ref{alg:IRRTs} and used in the subroutine $NoCollision(x_{nearest},x_{new})$. The bottom subfigure is the same as the top, but with the covariance tube as defined by \eqref{eq:prob_ellipse} overlaid, illustrating how the buffer prevents the tube from intersecting the true obstacle.

\begin{figure}[htbp!]
	\centering
	\includegraphics[width=0.5\linewidth]{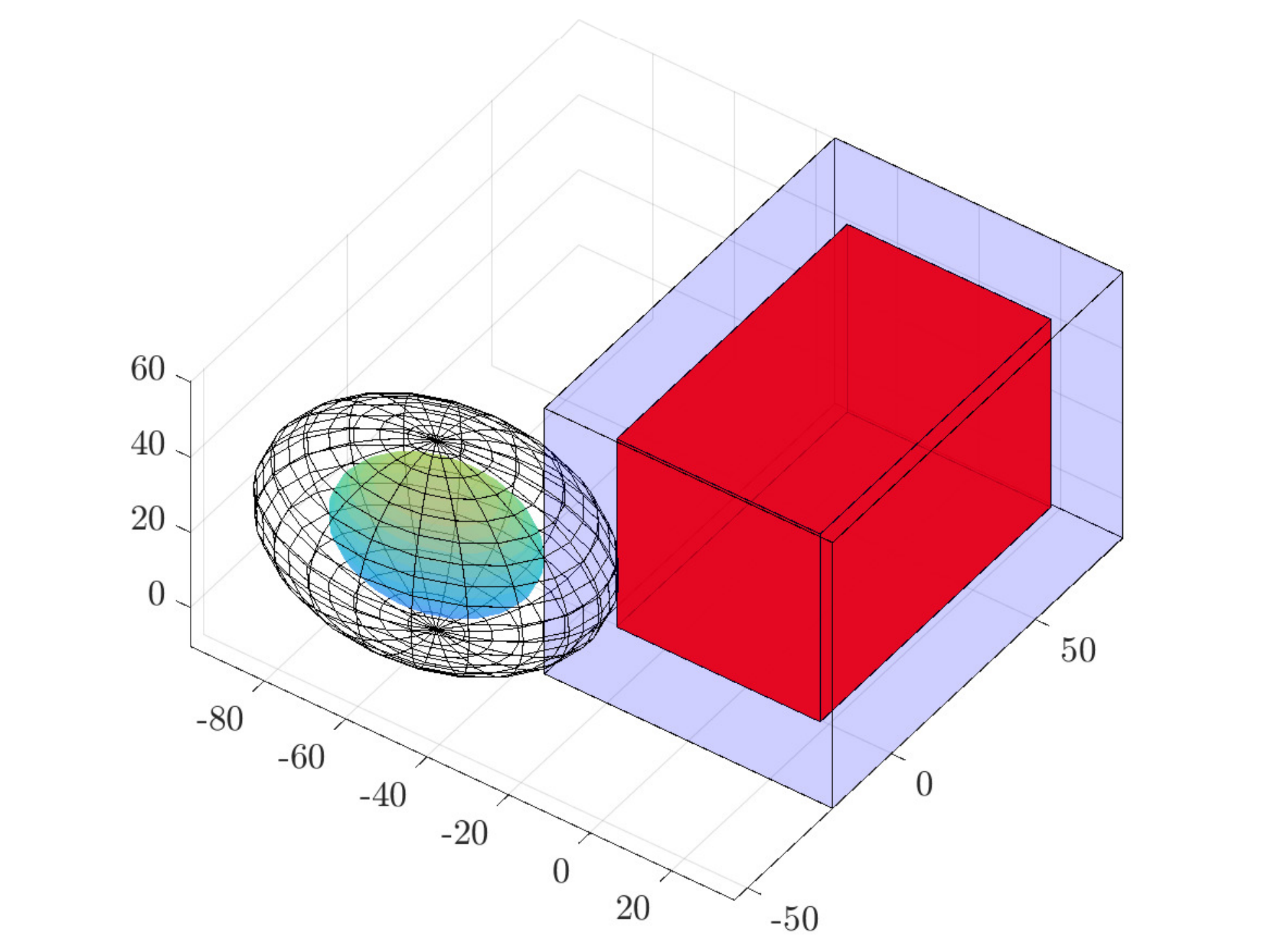}
	\caption{Cuboid buffer expansion illustration. The shaded ellipsoid and red cuboid represent the original confidence ellipsoid and obstacle, respectively. The wire mesh ellipsoid and blue cuboid represent the expanded sets that just touch the original obstacle and confidence ellipsoid, respectively.}
	\label{fig:cuboid_expansion}
\end{figure}

The dynamic extension to this algorithm involves intermittently halting the standard RRT algorithm, propagating the nominal vehicle trajectory and covariance, and adjusting the obstacle buffer sizes accordingly. This requires an algorithm that informs the buffer size adjustments, answering the question: "By how much should we grow or shrink the buffer such that the covariance tube just touches the obstacle?". To answer this, we start with the easier question of "What size ellipsoid just touches the obstacle?" This is illustrated in Figure \ref{fig:cuboid_expansion} and can be expressed as a QP: 

\begin{align}
&\min_{z} (z - \bar{r}(t))^{\top}\Sigma^{-1}_r(t) (z - \bar{r}(t)) \\
& s.t. \nonumber \\ 
& A_O z \leq b_O . \label{eq:qp_constraints}
\end{align}

We alter \eqref{eq:qp_constraints} to take into account the size of the buffer as follows:

\begin{align}
& A_O z \leq b_O + ld 
\end{align}

\noindent where $l$ is a vector of ones and $d$ is a scalar representing the size of the buffer. We can then vary $d$ until $c^{2*} = c^2$, which solves for the buffer size such that the smallest ellipsoid that intersects the buffer is the same size as the actual covariance ellipsoid. This solution is used as the buffer size for the next iteration. Note that this new buffer size is only an approximation of the actual solution since the size of the covariance tube is dependent on the solution to the inner loop, $X_{des}$, which changes every iteration. 

Letting $l$ be a vector of all ones has the effect of buffering the obstacle by an equal amount in all directions, as opposed to extending the buffer only in certain directions. Equal buffering on all sides of the obstacle may result in less optimal trajectories in a situation where the vehicle's trajectory interacts with multiple edges of an obstacle and this is viewed as the trade-off for not including the computationally nontrivial logic for a more discriminating buffer re-sizing. Additionally, expanding the buffer in all directions reduces the RRT search space more than merely expanding in one direction, which aids convergence.

Because the size of the probability ellipsoids vary spatially, the optimal buffer size is not necessarily the same for every obstacle. Thus, buffer changes happen independently for every obstacle. There are two cases to be considered: the buffer shrinks or stays the same, and the buffer grows. If a buffer is shrunken or left unchanged, no RRT nodes are changed, and the algorithm continues as normal. However, if a buffer grows, the nodes and connecting vertices that are now contained within the newly-sized obstacle are no longer valid. They are thus deleted, leaving only the parent trees and the now-stranded branches. Then, the Informed RRT* algorithm is used to grow branches from the parent tree until any part of the stranded branch is re-connected. At this point, any nodes left without a parent are removed from the set of nodes, and the Informed RRT* algorithm continues to run according to Algorithm \ref{alg:IRRTs} until the next covariance propagation and buffer resize. This is laid out in more detail in Algorithm \ref{alg:iRRTsD}.

\begin{algorithm}
	\caption{$\mathcal{T} = $ Dynamic Informed RRT$^*$($x_0, x_f$)}
	\label{alg:iRRTsD}
	\begin{algorithmic} [1]
		\State $\mathcal{T}_0 \leftarrow x_0 $
		\State $c_{best} \leftarrow \infty $
		\For{$m = 1 \dots M$}
		\State $k = 0$
		\While {$k < N_{max}$ AND $\delta > tol$}
		\State $k = k+1$
		\State $\mathcal{T}_{k} = AddNode(\mathcal{T}_{k-1})$
		\State $cost_i = FindBestPath(\mathcal{T}_{i})$
		\If {$i>N_{conv}$}
		\State $\delta = \lvert  \frac{cost_{i}-cost_{i-N_{conv}}}{cost_{i-N_{conv}}}  \rvert$
		\EndIf
		\EndWhile
		\If{$m \ne M$}
		\State $d = CompObsDist()$ \% Compute distance between obstacle and covariance tube
		\For{$j = 1 \dots N_{obs}$}
		\If{$d_j \geq 0$} 
		\State $b_j = b_j-  d_j$
		\ElsIf{$d_j <0$}
		\State $b_j = b_j-  d_j$
		\State $\mathcal{T} = CleanupNodes(\mathcal{T},O_j)$ \% remove nodes that are no longer valid
		\State $\mathcal{T} = Regrow(\mathcal{T},O_j,params)$ \% regrow until orphaned nodes are reconnected or pruned
		\EndIf
		\EndFor
		\EndIf
		\EndFor \\
		\Return{$\mathcal{T}$}
	\end{algorithmic}
\end{algorithm}

\section{Results} \label{results}

\subsection{Trajectory Validation}

The first result of interest is a validation of the linear covariance (LC) propagation described in Section \ref{sec:cov_prop} versus the results of the Monte Carlo simulations. For both the fixed-wing and quadrotor models, a simple desired trajectory was prescribed, and the state covariance computed from a 10,000 run MC simulation is compared to the LC results computed from Eqn. \eqref{eq:LinCov}. For the quadrotor model, the desired trajectory is a three phase ascent-cruise-descent mission profile. For the fixed-wing model, it is a constant-altitude trajectory with a sinusoid shape in the lateral direction. 

\begin{figure}[htbp!]
	\centering
	\includegraphics[width=0.5\linewidth]{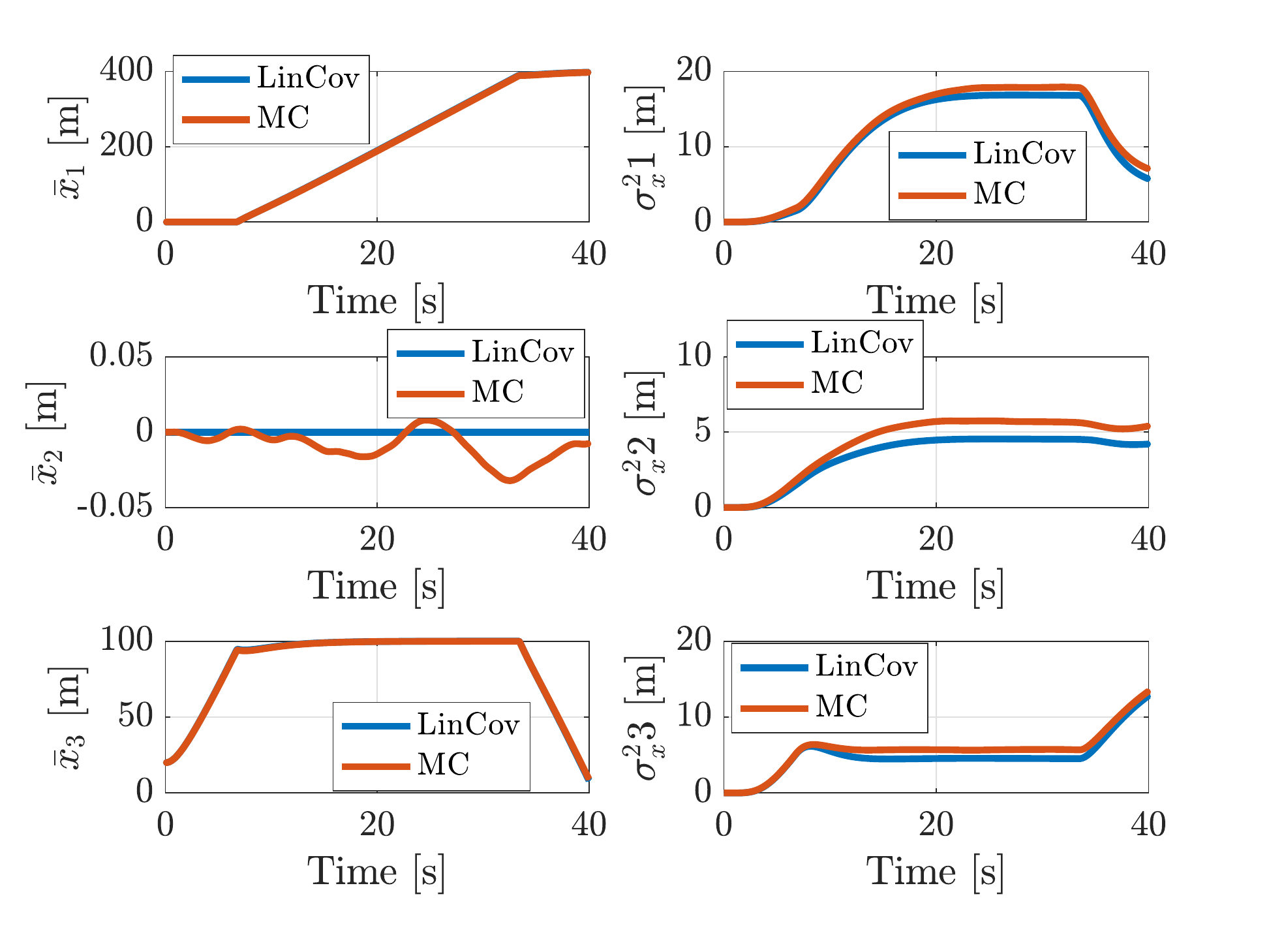}
	\caption{State and covariance time histories for prescribed quadrotor trajectory.}
	\label{fig:QR_MC_cov}
\end{figure}

\begin{figure}[htbp!]
	\centering
	\includegraphics[width=0.5\linewidth]{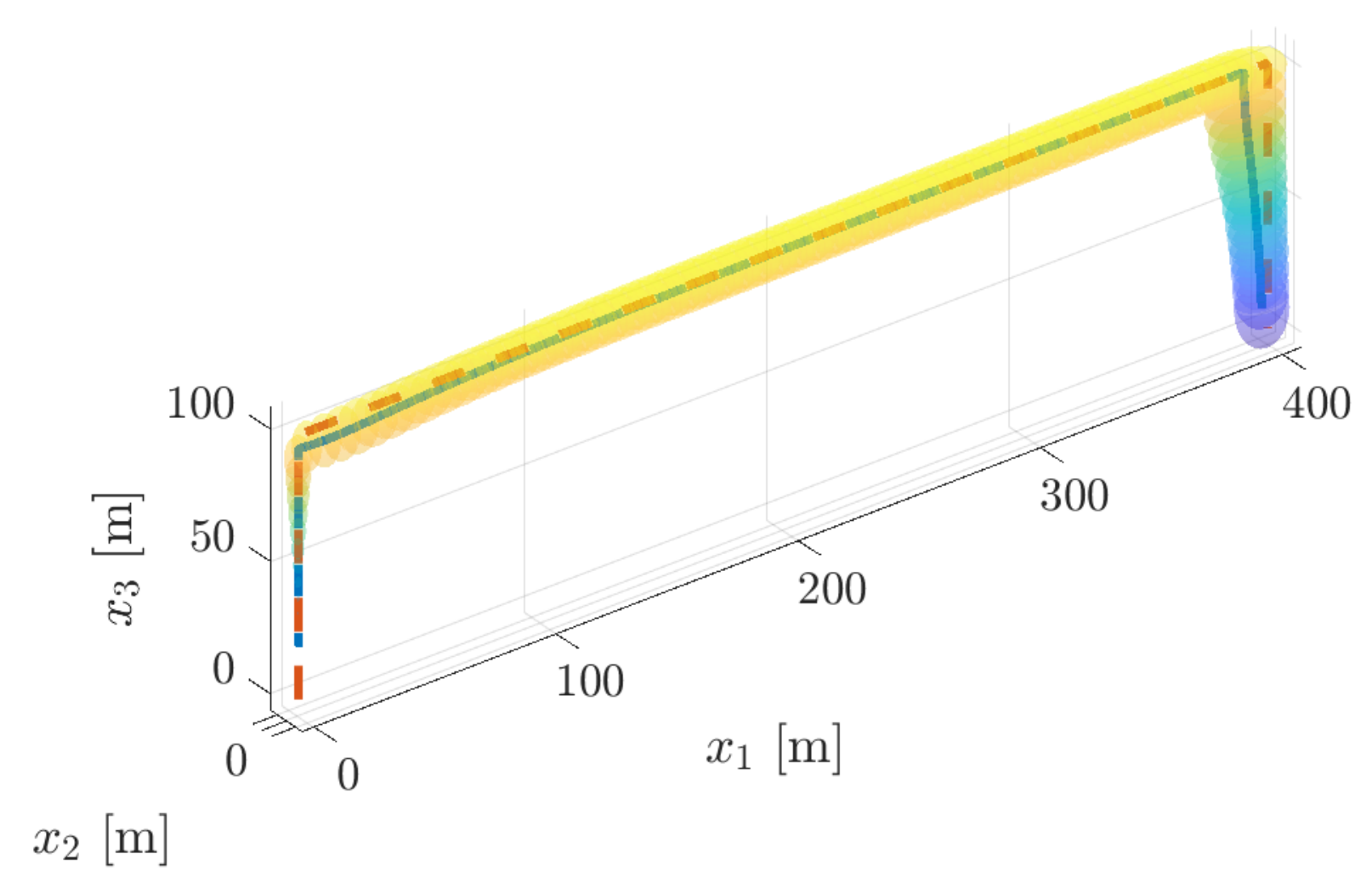}
	\caption{Quadrotor 3D trajectory. The covariance tube is constructed from a set of confidence ellipsoids defined by \eqref{eq:prob_ellipse}, the dashed line is the desired trajectory, and the solid line is the nominal trajectory $\bar{X}(t)$.}
	\label{fig:QR_MC_3D_traj}
\end{figure}

\begin{figure}[htbp!]
	\centering
	\includegraphics[width=0.5\linewidth]{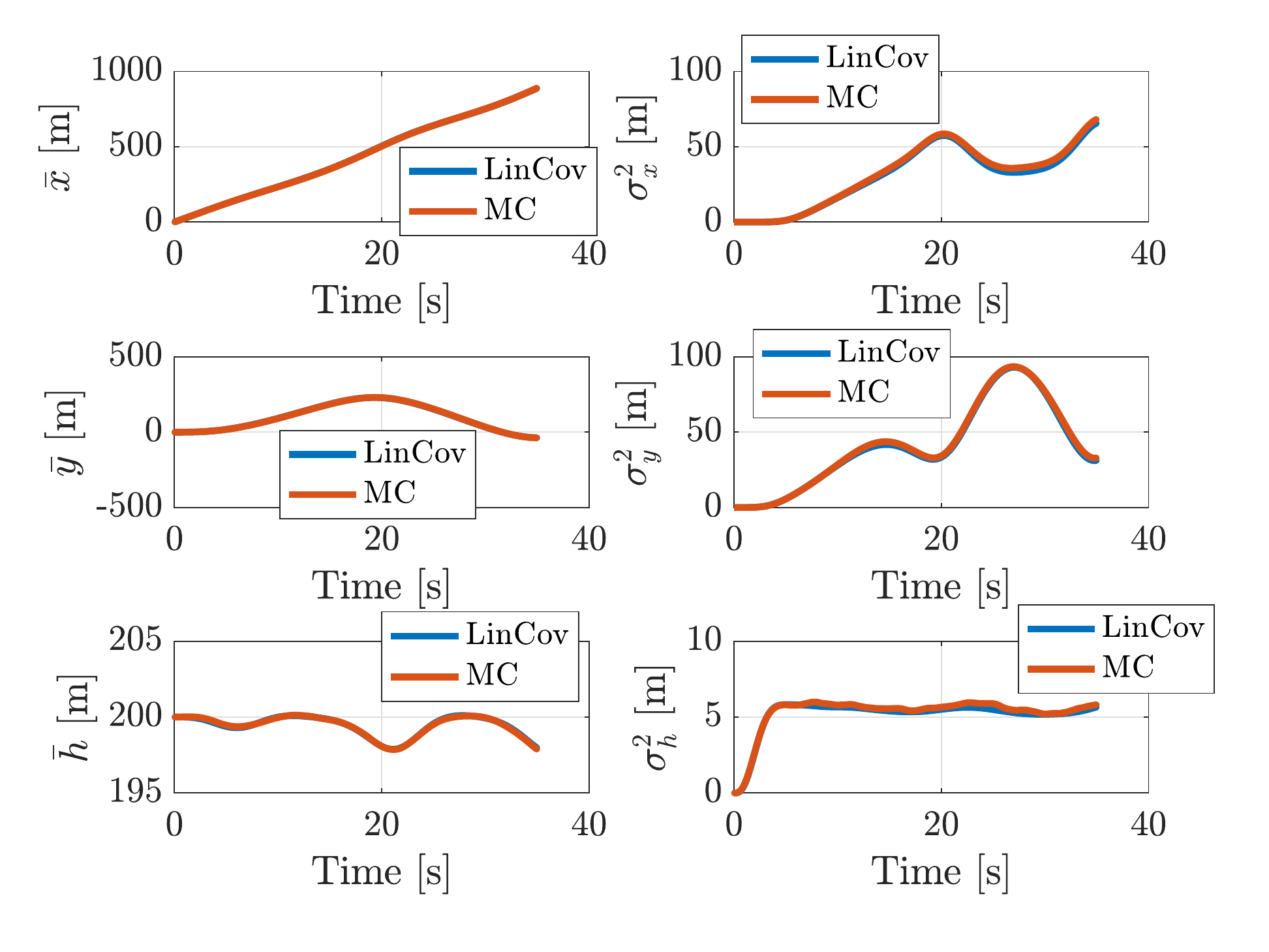}
	\caption{State and covariance time histories for prescribed fixed-wing trajectory.}
	\label{fig:FW_MC_cov}
\end{figure}

\begin{figure}[htbp!]
	\centering
	\includegraphics[width=0.5\linewidth]{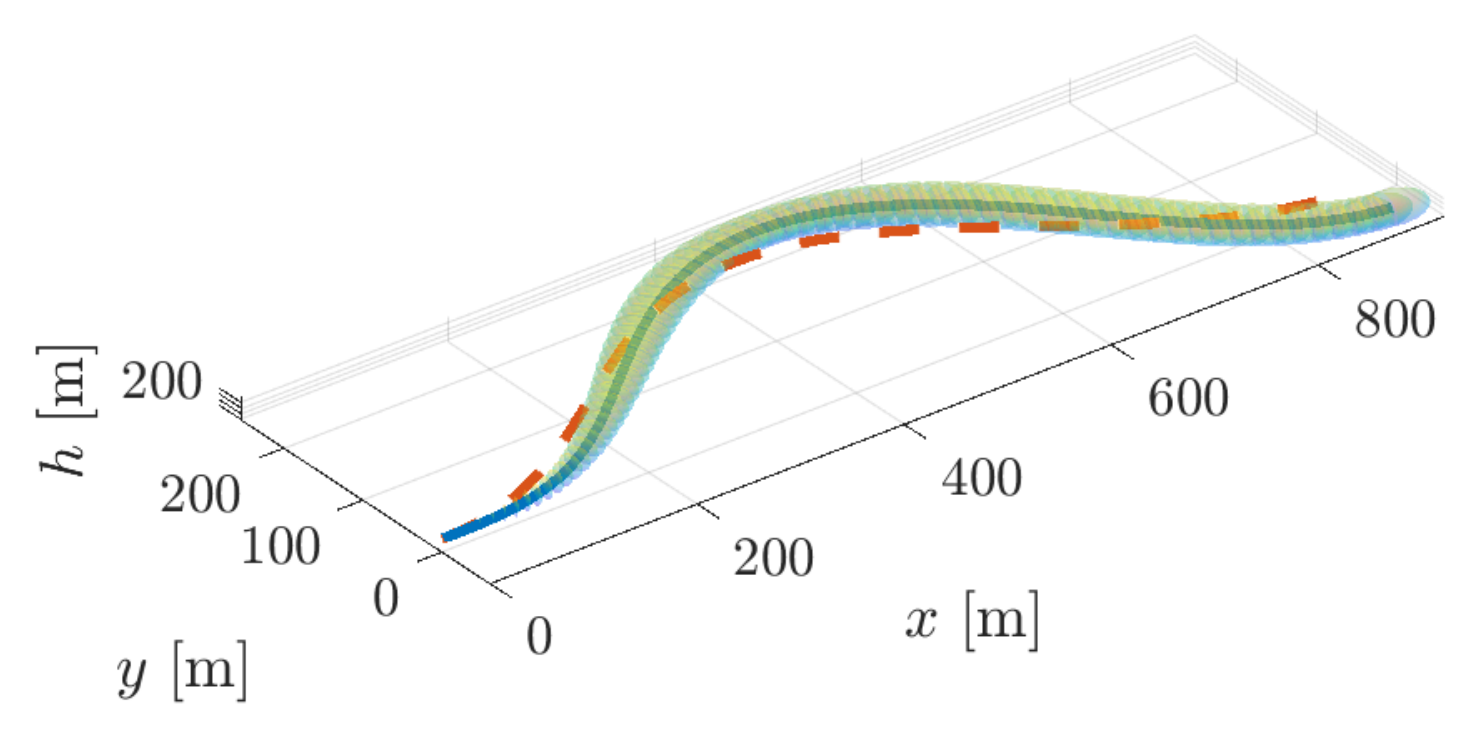}
	\caption{Fixed-wing 3D trajectory. The covariance tube is constructed from a set of confidence ellipsoids defined by \eqref{eq:prob_ellipse}, the dashed line is the desired trajectory, and the solid line is the nominal trajectory $\bar{X}(t)$.}
	\label{fig:FW_MC_3D_traj}
\end{figure}

The comparison results are shown in Figures \ref{fig:QR_MC_cov} -- \ref{fig:FW_MC_3D_traj}. Here, $\bar{x}_1$, $\bar{x}_2$, $\bar{x}_3$, $\bar{x}$, $\bar{y}$, and $\bar{h}$ represent either the mean position, or the nominal trajectory $\bar{X}(t)$, from the MC and LC simulations, respectively. All plots show good agreement between MC and LC, with the largest deviations being  $\approx 15\%$ in the
$\sigma_{x2}^2$ channel in Figure \ref{fig:QR_MC_cov}. This justifies our use of LC propagation in path planning under uncertainty. Note that the deviations in $\bar{x}_2$ appear large in the plot, but the vertical axis range is quite small. Also note that the nominal trajectory for the fixed-wing example does not track the desired trajectory exactly, but that is a function of the model's controller, which is not the focus of this work, as explained in Section \ref{sec:modeling}.

All computations are performed in the MATLAB environment running on a dual-core Intel Core i5 at 2.7 GHz, and all confidence ellipsoid tubes use $\beta = 0.999$. This value was chosen for illustration purposes, though some UTM applications may call for smaller or larger values. For the quadrotor model, the LC propagation took $0.062 s$ of computation time for $40 s$ of simulated flight time. For the fixed-wing model, it took $0.18 s$ of computation time for $35 s$ of simulated flight time. 

\subsection{Path Planning}

Now that we have shown that the LC propagation scheme is sufficiently fast and accurate when compared to MC simulations of the nonlinear system, we can use the LC propagation algorithm to enable chance-constrained path planning as laid out in Section \ref{sec:path_planning}. For brevity, only the quadrotor model is utilized in this section. 

\begin{figure}[htbp!]
	\centering
	\includegraphics[width=0.5\linewidth]{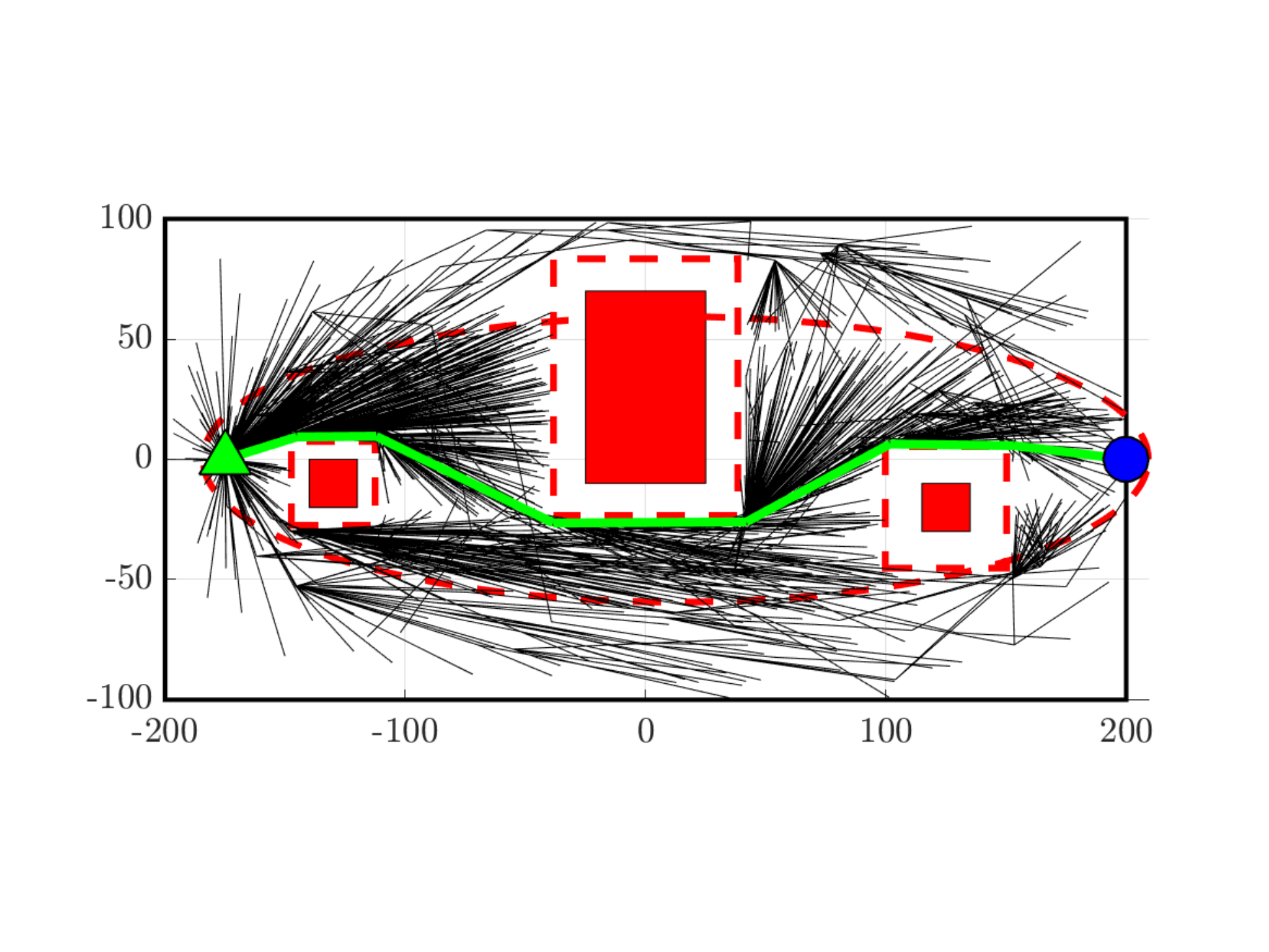}
	\caption{Chance-constrained Informed Dynamic RRT* solution for quadrotor model with three obstacles.}
	\label{fig:QR_RRT_nodes}
\end{figure}

\begin{figure}[htbp!]
	\centering
	\includegraphics[width=0.5\linewidth]{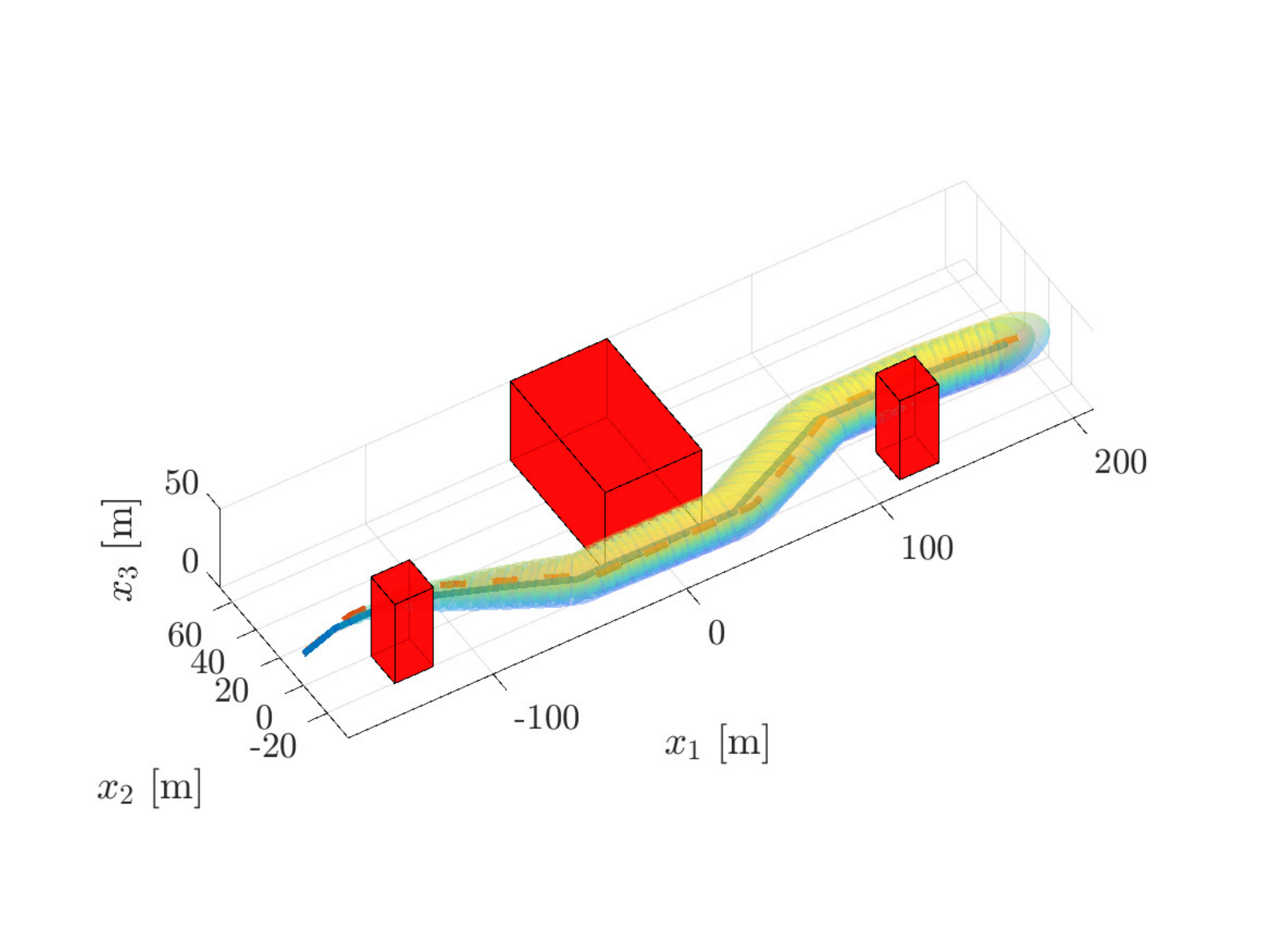}
	\caption{3D plot of obstacles and flight path with $\beta$ confidence tube overlaid.}
	\label{fig:QR_RRT_3D_tube}
\end{figure}

Figures \ref{fig:QR_RRT_nodes} and \ref{fig:QR_RRT_3D_tube} show the results of Algorithm \ref{alg:iRRTsD} for a three obstacle scenario. A solution is found that is very close to the minimum path length between $x_{start}$ and $x_{goal}$, while still ensuring that the $\beta$ confidence ellipsoid tube does not intersect any of the obstacles. Total computation time for this example was $30.5 s$. 

Note that the buffer re-sizing algorithm converges, but only to a region, the size of which is dictated by $tol$. There is no guarantee that the algorithm will exit without a constraint violation, but, based on extensive simulation studies, the magnitude of the violation will be sufficiently small for sufficiently large $M$ and small $tol$.

\section{Conclusions}

Future UTM systems will need to rely on rapid UQ for trajectory validation and path planning in order to account for the effects of wind turbulence/gusts and other uncertainties and disturbances. This work combines existing Dynamic RRT* and Informed RRT* algorithms, and adds an obstacle buffer resizing technique to solve a challenge of chance-constrained path planning: trajectory re-planning changes the outcome of the covariance propagation. The results presented here shows that this RRT*-based algorithm, in combination with QP-based collision checking for trajectory validation, successfully solves the aforementioned problem, resulting in a computationally efficient chance-constrained path planner. Trajectory validation examples were presented for both quadrotor and fixed-wing models in 3D flight in a non-static atmosphere. Path planning examples were presented for the quadrotor model, showing near-optimal navigation around 3 obstacles while enforcing chance state constraints.

\subsection*{Financial disclosure}

This work was supported in part by NASA under Cooperative Agreement NNX16AH81A.

\subsection*{Conflict of interest}

The authors declare no potential conflict of interests.

\appendix

\section{Quadrotor Modeling} \label{app:quad_model}

As a balance between model fidelity and computational efficiency, the quadrotor model used for uncertainty propagation is built upon double integrator dynamics augmented with aerodynamic drag of the following form:

\begin{align}
& {r} = 
\begin{bmatrix}
x_1 \\
x_2 \\
x_3
\end{bmatrix} , ~
{V}_0 = 
\begin{bmatrix}
\dot{x}_1 \\
\dot{x}_2 \\
\dot{x}_3 
\end{bmatrix}, ~
{V_q}={V}_0 - {w}, \\ \label{eq:quad_dr}
& \dot{{V}}_0 = {u} - \frac{1}{2m} \rho S C_D {V_q} \left\lVert {V_q}  \right\rVert ,
\end{align}

\noindent where $x_1$, $x_2$, and $x_3$ are the vehicle's coordinates in an inertial reference frame, ${V_q}$ is its velocity with respect to the atmosphere, ${w} \in \mathbb{R}^3$ is the velocity of the local atmosphere with respect to the ground, ${u} \in \mathbb{R}^3$ is the control input, $m$ is the vehicle mass, $\rho$ is the air density, $S$ is the reference area, and $C_D$ is the coefficient of drag, assumed to be constant. This model assumes that an inner-loop controller for vehicle attitude and thrust has been implemented that has a feed forward term to cancel out gravity and has sufficiently high bandwidth that we can control acceleration directly. It also assumes a non-static atmosphere and a drag coefficient that remains constant regardless of vehicle state or direction of the relative wind vector ${V_q}$. Uncertainty in the drag coefficient can, however, be handled using our UQ and planning algorithms.

When using the above model for uncertainty propagation, it is important that a controller be added so that the uncertainty in vehicle states does not grow so large as to be useless for trajectory prediction purposes. The method presented here is agnostic to controller choice, but for modeling purposes, it is beneficial if the controller can be expressed in a closed form that is easily linearizable. This work utilizes a dynamic extension controller of the following form: 

\begin{align}
{e}  &=~ r - r_{des}, \\
{S_c} &=~ \dot{{e}} + K_q e,\\
{u} &=~ \ddot{{r}}_{des}-K_q \dot{{e}} - \Lambda_q {S_c},
\end{align} 

\noindent where $K_q \in \mathbb{R}^{3 \times 3}$ and $\Lambda_q \in \mathbb{R}^{3 \times 3}$ are controller gains that may be treated as tuning parameters to match the flight characteristics of our model to the flight characteristics of a real-world vehicle for which the UTM may be trying to predict its trajectory. 

Finally, we consider the model for our wind disturbance. A Dryden wind model specifies the power spectral density (PSD) for the body-fixed longitudinal, lateral, and vertical directions of a fixed-wing aircraft. Other works \cite{waslander2009wind} apply this model directly to quadrotors. Here, the longitudinal channel is replicated in the $x_1$, $x_2$, and $x_3$ directions of the inertial frame. The resulting Dryden-like wind model is summarized as

\begin{align}
H_i(s) &=~ \sigma_i \sqrt{\frac{2L_i}{\pi ||V_0||}} \frac{1}{1+ \frac{L_i}{||V_0||}s} ,\\
A_i &=~ \frac{-||V_0||}{L_i}, \\
B_i &=~ 1, \\
C_i &=~ \sqrt{2} ||V_0|| \sigma_i \sqrt{\frac{L_i}{||V_0||}} \frac{1}{L_i \sqrt{\pi}} ,\\
\dot{\eta}_i &=~  A_i \eta_i + B_i n, \\
w_i &=~ C_i \eta_i,  \label{eq:quad_wi}
\end{align}

\noindent for $i = \{x_1,x_2, x_3\}$. Here, $H_i$ is the transfer function of  Dryden model coloring filter, $A_i$, $B_i$, and $C_i$ are its state space realization, $\sigma_i$ is the gust intensity parameter, $L_i$ is the characteristic length, and $n$ is the Gaussian white noise input \cite{hoblit1988gust}. The output of this model, ${w}=[w_1,w_2,w_3]^\top$, is the wind velocity vector in (\ref{eq:quad_dr}). 

\section{Fixed-Wing Aircraft Modeling} \label{app:fixed_model}

To illustrate another typical setting which involves sUAS, the EOMs for a fixed-wing aircraft under closed-loop control in a non-static atmosphere were derived. Hull \cite{hull2007fundamentals} provides EOM's for longitudinal flight with moving atmosphere and EOM's for full three-dimensional flight in a static atmosphere, but does not provide EOM's for 3D flight in a moving atmosphere. 

\begin{figure}[htbp!]
	\centering
	\includegraphics[width=4in]{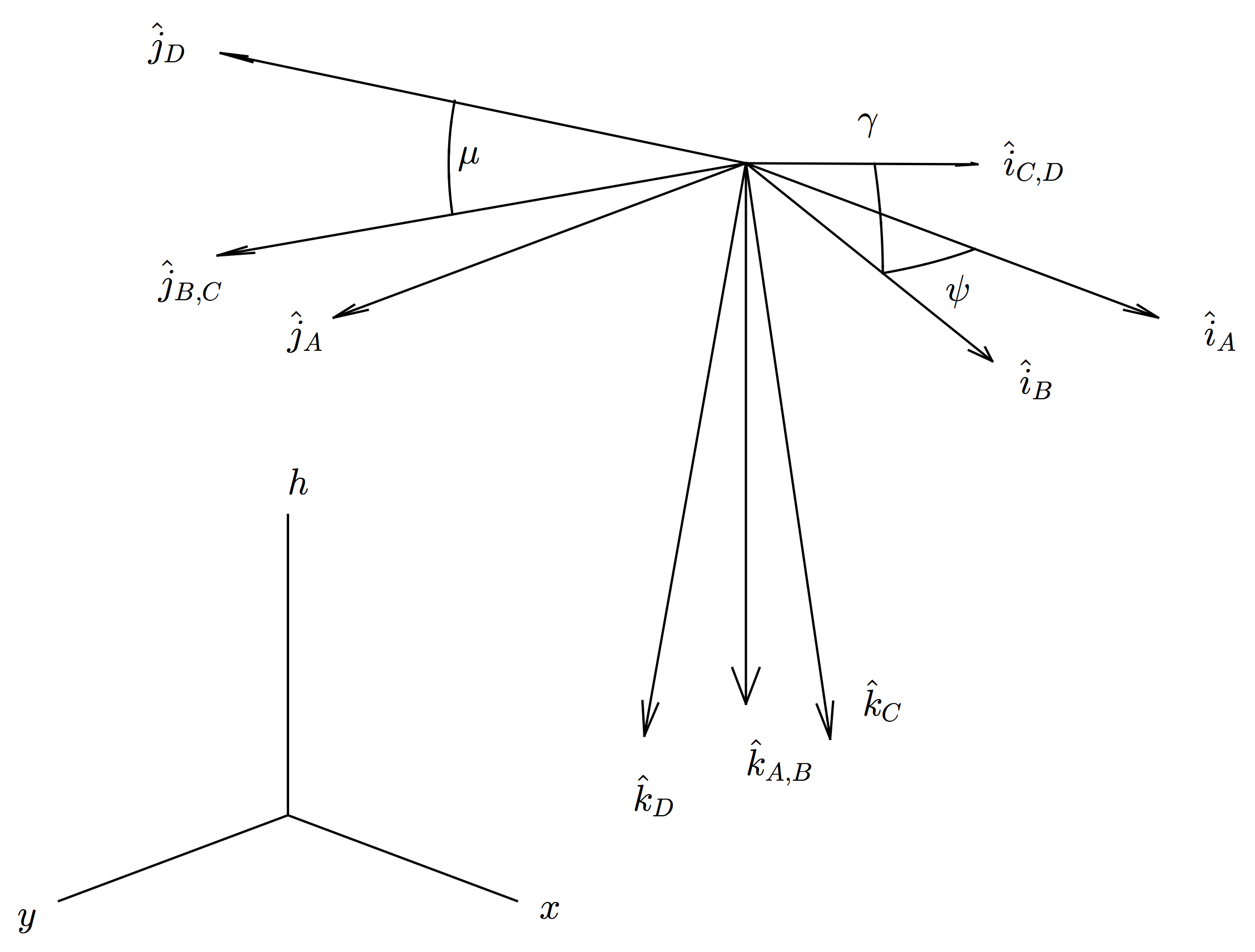}
	\caption{Three-dimensional flight: unit vectors and rotations.}
	\label{fig:3D_unit_vec}
\end{figure}

Figure \ref{fig:3D_unit_vec} shows the unit vectors and rotations necessary to define flight in three dimensions. Here, frame $A$ is the inertial frame, while frames $B$, $C$, and $D$ are rotated by angles $\psi$, $\gamma$, and $\mu$, respectively, where $\psi$ is velocity yaw or heading angle, $\gamma$ is velocity pitch, and $\mu$ is velocity roll. 

\begin{figure}[htbp!]
	\centering
	\includegraphics[width=4in]{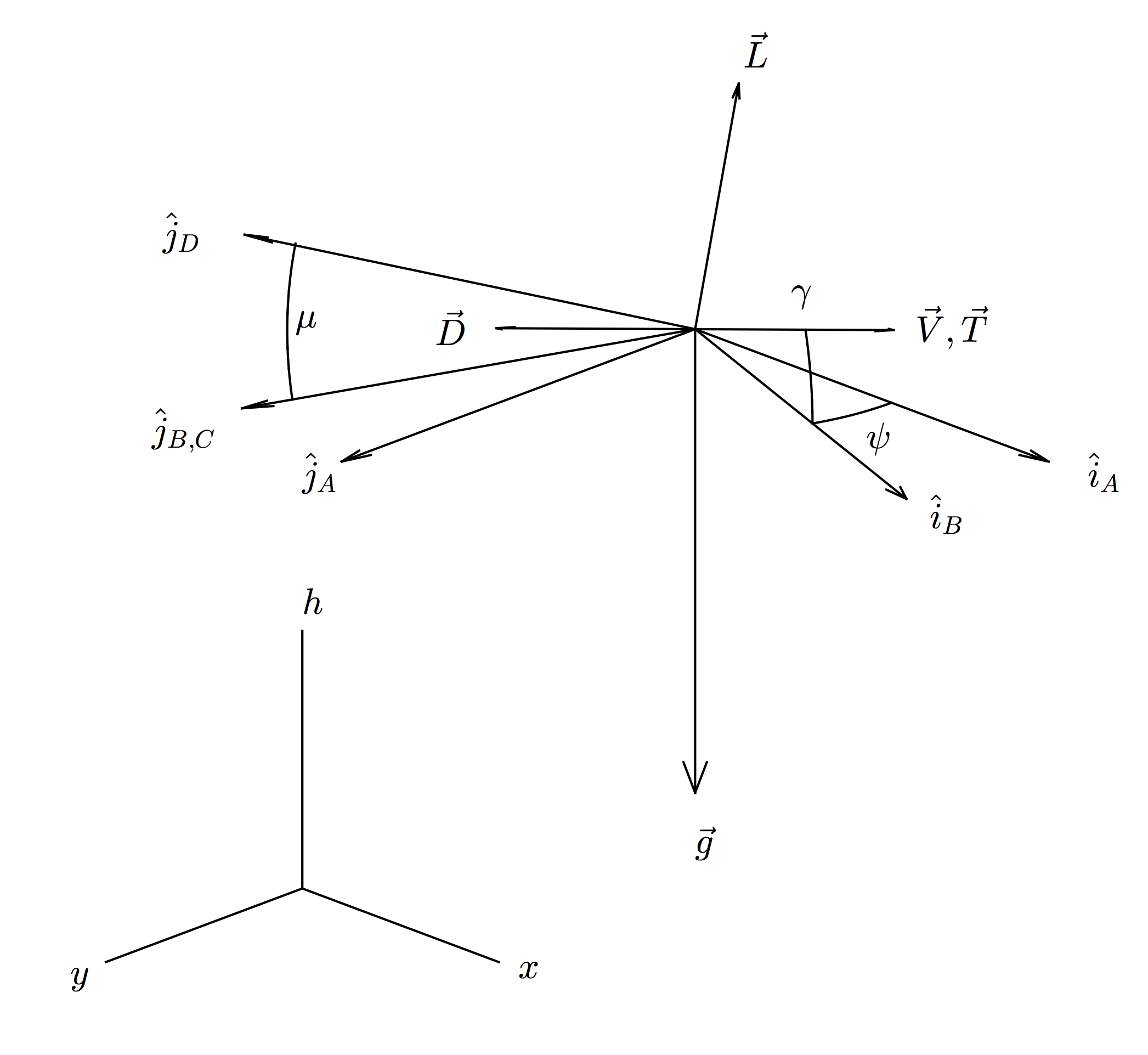}
	\caption{Three-dimensional flight: forces and velocity.}
	\label{fig:3D_forces}
\end{figure}

Additionally, the thrust, drag, lift and gravity 
forces acting on the aircraft can be defined as follows:

\begin{align}
\vec{T}=T\hat{i}_D, \\ \label{eq:fixedw_T}
\vec{D}=-D\hat{i}_D, \\
\vec{L} = -L\hat{k}_D ,\\
m\vec{g}=mg\hat{k}_A,
\end{align}

\noindent and the velocity of the aircraft can be represented as $\vec{V} = V\hat{i}_D$, as seen in Figure \ref{fig:3D_forces}. 

following equations:  

\begin{align}
\mu = &~ \kappa_{\mu}(\psi_{des}-\psi), \\
C_L = &~ \bar{C_L} + \kappa_{C_L} (\gamma_{des} - \gamma), \\
\dot{T} = &~ \kappa_{T,1} (T_{des}-T), \\
T_{des} = &~ \bar{T} + \kappa_{T,2} (V_{des} - V).
\end{align}

\noindent Here, $\psi_{des}$, $\gamma_{des}$, $T_{des}$, and $V_{des}$ are the desired values of the vehicle's yaw, pitch, thrust, and velocity, respectively. The feed forward terms $\bar{C_L}$ and $\bar{T}$ are the values of $C_L$ and $T$, respectively, that are required to maintain steady flight: 

\begin{align}
\bar{C_L} = &~ \frac{2mg\cos(\gamma)}{SV^2 \rho}, \\
\bar{T} = &~ mg\sin (\gamma) + \frac{1}{2}C_{D0}SV^2 \rho + \frac{2 K_{d} m^2 g^2 \cos (\gamma)^2}{SV^2 \rho}.
\end{align}

The vehicle's drag polar parameter is denoted $K_d$. Similar to the quadrotor model, controller gains $\kappa_{\mu}$, $\kappa_{C_L}$, $ \kappa_{T,1}$, and $ \kappa_{T,2}$ may be tuned to better replicate the behavior of a real-world UAS. 

The outer-loop controller is separated into two components: longitudinal and lateral. Its design requires us to add two additional states ($V_{des}$ and $\psi_{des}$) and to estimate $\ddot{\eta}_{des}$ via finite differences, which is undesirable from a computational efficiency perspective, but avoids wrap-around issues. 

For the longitudinal controller,  $V$ is assumed to be constant and only the $\dot{h}$ dynamics are considered:

\begin{align}
&\dot{h} = V \sin(\gamma), \\ 
&\mathrm{Control:} ~\gamma, \\ 
& e = h-h_{des}, \\
& \dot{e} = -\kappa e, \\
& \gamma_{des} = \sin^{-1}\bigg ( \frac{\dot{h}_{des}-\kappa (h-h_{des})}{V} \bigg ).
\end{align}

Note that because above controller is part of the outer-loop, the output is $\gamma_{des}$ and not $\gamma$. 

For the lateral controller, we need to control $\dot{\psi}$ so a different approach is pursued. Here, $\gamma$ is assumed to be constant and only the $\dot{x}$ and $\dot{y}$ kinematics are considered:

\begin{align}
\dot{x} =&~ V\cos(\gamma)\cos(\psi), \\
 \dot{y} = &~V\cos(\gamma)\sin(\psi) ,\\
&\mathrm{Controls: } ~\dot{V}, \dot{\psi} ,\\ 
\eta = &~
\begin{bmatrix}
x \\
y
\end{bmatrix},~ e = \eta - \eta_{des} ,\\
 S =&~ \dot{e}+\kappa e ,\\
 \dot{S} =&~ \ddot{\eta} - \ddot{\eta}_{des} + \kappa \dot{e}, \\
 \dot{S} = &~
\begin{bmatrix}
\cos(\gamma) \cos(\psi_{des}) & -V_{des}\cos(\gamma)\sin(\psi_{des}) \\ 
\cos(\gamma) \sin(\psi_{des}) & V_{des} \cos(\gamma) \cos(\psi_{des})
\end{bmatrix}
\begin{bmatrix}
\dot{V}_{des} \\
\dot{\psi}_{des}
\end{bmatrix} \nonumber \\
&- \ddot{\eta}_{des} + \kappa \dot{e} ,\\
 \dot{S} =&~ A \nu - \ddot{\eta}_{des} + \kappa \dot{e} ,\\
 \dot{S} =&~ -\Lambda_f S, ~ -\Lambda_f ~ \mathrm{Hurwitz}, \\
\begin{bmatrix}
\dot{V} \\ \dot{\psi}
\end{bmatrix}= &~A^{-1}[ \ddot{\eta}_{des} - \kappa \dot{e} - \Lambda_f S ].
\end{align}

This control is invalid if $A$ is singular, corresponding to a pitch angle of $\pm 90^\circ$ or a desired velocity of $V_{des}=0$, which are flight conditions not encountered in this work. 

For the wind disturbance, the Dryden wind model defines wind in the longitudinal, lateral, and vertical body fixed directions,  \textit{not} in the $x, y, h$ directions. In the two dimensional case \cite{berning2018Rapid}, we assumed that $\gamma$ is small and thus we have wind in the $x$ and $h$ directions. In the three-dimensional case, however, we cannot necessarily make the same assumptions about $\psi$. Thus, the following wind definitions are used:

\begin{align}
{w} & =w_u \hat{i}_D+w_w \hat{j}_D +w_v \hat{k}_D  = w_x \hat{i}_A+w_y \hat{j}_A + w_h \hat{k}_A.
\end{align}

We can now write the full equations of motion as follows:

\begin{align}
\dot{x} = &V \cos(\gamma) \cos(\psi) +w_x \label{fixed_dx},\\ 
\dot{y} =& V \cos(\gamma) \sin(\psi) +w_y, \\
\dot{h} = &V \sin(\gamma)+w_h, \\
\dot{V} =& \frac{T-D}{m}-g\sin(\gamma)-\dot{w}_x \cos(\gamma) \cos(\psi) \nonumber \\ 
&-\dot{w}_y \cos(\gamma) \sin(\psi) + \dot{w}_h \sin(\gamma)  \label{dV},\\
\dot{\psi} = &\frac{-1}{Vm\cos(\gamma)} \big [ L\sin(\mu)-m\dot{w}_x \sin(\psi) \nonumber \\ 
& + m \dot{w}_y \cos(\psi)  \big ] \label{dpsi},\\
\dot{\gamma} =& \frac{1}{Vm} \big [ L \cos(\mu) -mg\cos(\gamma) + m\dot{w}_x \cos(\psi)\sin(\gamma) \nonumber \\ 
& + m\dot{w}_y \sin(\gamma) \sin(\psi) + m \dot{w}_h \cos(\gamma) \big ] \label{dgamma},
\end{align}

\noindent where

\begin{align}
w_x = &~w_u\cos(\gamma)\cos(\psi) - w_w(\cos(\mu)\sin(\psi) \nonumber \\ 
&+ \cos(\psi)\sin(\gamma)\sin(\mu)) - w_v(\sin(\mu)\sin(\psi) \nonumber \\ 
& - \cos(\mu)\cos(\psi)\sin(\gamma)) , \\
w_y = &~w_v(\cos(\psi)\sin(\mu) + \cos(\mu)\sin(\gamma)\sin(\psi)) \nonumber \\
& + w_w(\cos(\mu)\cos(\psi) - \sin(\gamma)\sin(\mu)\sin(\psi)) \nonumber \\ 
& + w_u\cos(\gamma)\sin(\psi) ,\\
w_h = &~w_v\cos(\gamma)\cos(\mu) - w_u\sin(\gamma) \nonumber \\ 
& - w_w\cos(\gamma)\sin(\mu),
\end{align}

\begin{align}
C_D =&~ C_{D0}+K_{d} C_L^2 \label{eq:drag_polar} ,\\ 
L=&~\frac{1}{2}C_L\rho SV^2 ,\\
D=&~\frac{1}{2}C_D\rho SV^2 . \label{eq:drag}
\end{align}

If we assume that $ \lvert \dot{w}_u \rvert,\lvert \dot{w}_w \rvert,\lvert \dot{w}_v \rvert \gg \lvert \dot{\gamma} \rvert, \lvert \dot{\psi} \rvert, \lvert \dot{\mu} \rvert$ then we can express the wind accelerations as follows:

\begin{align}
\dot{w_x} =& ~\dot{w}_u\cos(\gamma)\cos(\psi) - \dot{w}_w(\cos(\mu)\sin(\psi) \nonumber \\ 
& + \cos(\psi)\sin(\gamma)\sin(\mu))  - \dot{w}_v(\sin(\mu)\sin(\psi) \nonumber \\ 
& - \cos(\mu)\cos(\psi)\sin(\gamma)), \\
\dot{w_y} = &~\dot{w}_v(\cos(\psi)\sin(\mu) + \cos(\mu)\sin(\gamma)\sin(\psi)) \nonumber \\ 
& + \dot{w}_w(\cos(\mu)\cos(\psi) - \sin(\gamma)\sin(\mu)\sin(\psi)) \nonumber \\ 
&+ \dot{w}_u\cos(\gamma)\sin(\psi), \\
\dot{w_h} = &~\dot{w}_v\cos(\gamma)\cos(\mu) - \dot{w}_u\sin(\gamma) \nonumber \\ 
& - \dot{w}_w\cos(\gamma)\sin(\mu).
\end{align}

The power spectral densities (PSD) 
for the Dryden model are defined following \cite{hoblit1988gust}. 

For the longitudinal channel:
\begin{align}
\Phi_u (\Omega) = \sigma_u^2 \frac{2L_u}{\pi} \frac{1}{[1+L_u^2+\Omega^2]^2},
\end{align}

For the lateral channel:
\begin{align}
\Phi_w (\Omega) = \sigma_w^2 \frac{L_w}{\pi} \frac{ 1 + 3L_w^2 \Omega^2}{ [1+L_w^2 \Omega^2]^2},
\end{align}

For the vertical channel:
\begin{align}
\Phi_v (\Omega) = \sigma_v^2 \frac{L_v}{\pi} \frac{ 1 + 3L_v^2 \Omega^2}{ [1+L_v^2 \Omega^2]^2},
\end{align}

\noindent where $\Omega = \frac{\omega}{V}$. The corresponding coloring filters are as follows: 

\begin{align}
G_u(s) = &~\sigma_u \sqrt{\frac{2 L_u}{\pi V}}\frac{1}{1+\frac{L_u}{V}s} ,\\
G_w(s) = &~\sigma_w \sqrt{\frac{L_w}{\pi V}} \frac{1+\frac{\sqrt{3}L_w}{V}s}{ \big ( 1+\frac{L_w}{V}s \big ) ^2} ,\\
G_v(s) = &~\sigma_v \sqrt{\frac{L_v}{\pi V}} \frac{1+\frac{\sqrt{3}L_v}{V}s}{ \big ( 1+\frac{L_v}{V}s \big ) ^2}.
\end{align}

These have state space realizations of the form:

\begin{align}
\dot{\eta}_i = &~A_i \eta_i+B_i n ,\\
w_i = &~C_i \eta_i ,\\
\dot{w}_i = &~C_i A_i \eta_i + C_i B_i n ,
\end{align}

\noindent for $i \in \{u, w, v\}$, where $n$ is the Gaussian white noise input and 
the matrices are defined as follows: 

\begin{align}
&A_u = \frac{-V}{L_u}, A_w = \begin{bmatrix}
\frac{-2V}{L_w} & \frac{-V^2}{L_w^2} \\
1 & 0
\end{bmatrix},A_v= \begin{bmatrix}
\frac{-2V}{L_v} & \frac{-V^2}{L_v^2} \\
1 & 0
\end{bmatrix} ,\\
&B_u = 1,B_w = \begin{bmatrix}1 \\ 0 \end{bmatrix},B_v = \begin{bmatrix}1 \\ 0 \end{bmatrix}, \\
&C_u = \frac{\sqrt{2}V \sigma_u \sqrt{\frac{L_u}{V}}}{L_u \sqrt{\pi}}, C_w = \frac{V \sigma_w \sqrt{\frac{L_w}{V}}}{L_w \sqrt{\pi}}
\begin{bmatrix}
\sqrt{3} & \frac{V}{L_w}
\end{bmatrix}, \nonumber \\ 
& C_v = \frac{V \sigma_v \sqrt{\frac{L_v}{V}}}{L_v \sqrt{\pi}}
\begin{bmatrix}
\sqrt{3} & \frac{V}{L_v}
\end{bmatrix} \label{eq:fixedw_wC}.
\end{align}

%\nocite{*}% Show all bib entries - both cited and uncited; comment this line to view only cited bib entries;
\bibliography{main}%

\end{document}